# LEXICAL BASE AS A COMPRESSED LANGUAGE MODEL OF THE WORLD
## (on the material of the Ukrainian language)

Solomiya Buk
Department for General Linguistics, Ivan Franko National University of Lviv,
1 Universytetska St., Lviv, UA–79000, Ukraine

Summary
In the article the fact is verified that the list of words selected by formal statistical methods (frequency and functional genre unrestrictedness) is not a conglomerate of non-related words. It creates a system of interrelated items and it can be named "lexical base of language". This selected list of words covers all the spheres of human activities. To verify this statement the invariant synoptical scheme common for ideographic dictionaries of different language was determined.

**The selection principles of the Ukrainian language lexical base**

In Ukrainian linguistic studies dealing with modern lexical stratification researchers investigate the lexical groups differing stylistically, by time and by territory or by environment of their functioning. The word's stratum with the highest usage and, according to V. Moskovich [*Moskovich* 1969, p. 23–51], respectively, with the highest information density and the importance for the text understanding, was not the single research object. Such a lexical base separation, it detailed analysis in side of words composition and in side of its classification into the paradigmatic groups can demonstrate the answer to the question about the language system. The paradigmatic groups selection on this base, ascertaining different semantic relations between those groups, observing its semantic description in one language explanatory dictionaries will make easier the work on their adequate semantization.

The lexical base separation has the real theoretical foundation: one can consider the existence of the kernel vocabulary in any language as one of the universal feature in human lexicon organization" [*Serebrennikov* et al, p. 121].

Practically all the developed languages have such a lexical base, for example, English [*Elridge* 1911; *Thorndike* 1931; *Ogden* 1937; *Palmer* 1937; *West* 1953], German [*Hauch* 1931], Spanish [*Keniston* 1929], French [*Goudenheim et al* 1956], Polish [*Kurzowa & Zgolkowa* 1992], Russian [*Denisov* 1972; *Morkovkin* 1984], etc. The comparison between lexical bases of different languages also exists [*Eaton* 1934]. Practically, all those authors (except [*Ogden* 1937]) select their lists using statistical criteria from the frequency dictionaries, some of them take into consideration word's occurrence in different type of text. Taking into account previous world experience, we worked out our own techniques of the language base selection.

As far as we know, there are no special researches devoted to the quantitative correlation of the functional genres in the daily life speech of an average man. There are many controversies in the point of proportioning and choosing the whole language frequency dictionary size in practice. Large frequency dictionaries are built on the basis of different proportions of genres. From this point of view we try to compare some frequency dictionaries of French [*Juilland et al* 1970], Finnish (the information for Finnish language is from [*Tuldava* 1987, p. 56]), Slovenic [*Mistrík* 1969], Polish [*Saloni* 1990], and Russian [*Zasorina* 1977], [*Shtejnfeld* 1963] languages. The results are shown in Table 1.

*Table 1.*

| Dictionary \ Function. genre | FD French, % | FD Finnish, % | FD Slovenian, % | FD Polish, % | FD Russian [Zasorina 1977], % | FD Russian [Shtejnfeld 1963] % |
|---|---|---|---|---|---|---|
| belles-lettres | 20 | 11.5 | 3 0.2 | 20 | 25 | 12.5 |
| essay | 20 | | | | | |
| drama | 20 | | | 20 | 25 | 12.5 |
| poetry | | | 13.2 | | | |
| dialogue | | | 10.5 | | | |
| radio-program | | 9.2 | | | | 25 |
| journalistic | 20 | 26 | 14.6 | 40 | 25 | 25 |
| scientific | | | 31.5 | 20 | 25 | |
| literature for children | | | | | | 25 |
| different | | 43.3 | | | | |
| common corpus, word occurrences | 500 000 | 400 000 | 1 000 000 | 500 000 | 1 000 000 | 400 000 |

As it can be seen from the table, all the dictionaries consider belles-lettres and journalistic genres, four of them (French, Polish and both Russian) consider drama as the equivalent of spoken language, three of them (Slovenian, Polish and Russian Zasorina) take scientific texts into consideration. The fact of official genre lack attracts some attention. Certainly, it is somewhat presented in the newspaper and magazine language, but it cannot be confined by it. In order to select the lexical base, we decided to compare frequency dictionaries of five functional genres (due to the standard classification): belles-lettres, journalistic, colloquial (spoken language), scientific and official genres.

For Ukrainian language, there are only two frequency dictionaries: belles-lettres [*Perebyjnis* 1981] and journalistic [*Darchuk & Grjaznuhina* 1996], the principles of their building are quite similar. Three other of them (colloquial, scientific and official) were built by the author of this article [*Buk* 2003a], [*Buk* 2003b].

Aiming all the functional style corpora under consideration to be equivalent, we used the corpus size of 300 000 word occurrences for each of three our dictionaries, according to the corpus size of the journalistic genre frequency dictionary.

For the further appropriateness of those frequency dictionaries comparing, their building principles were equal as described in [*Darchuk & Grjaznuhina* 1996]. Our original frequency dictionaries comparing methods, which takes into account statistical methods and world experience, is described in [*Buk* 2004]. In particular, it takes into account the text coverage analysis.

The special program was written for such a frequency dictionaries comparison. It throws together all the dictionaries words in one (first) column named "word", in the next columns (they are indexed by the numbers of five frequency dictionaries) the every word frequency is fixed. The last column shows the word sum for all the dictionaries (see Table 2).

| word | 1 | 2 | 3 | 4 | 5 | sum |
|---|---|---|---|---|---|---|
| НОВИЙ | 262 | 155 | 495 | 434 | 179 | 1525 |
| ТОМУ | 206 | 444 | 296 | 379 | 171 | 1496 |
| ОРГАНІЗАЦІЯ | 14 | 12 | 460 | 205 | 745 | 1436 |
| МОЖНА | 262 | 419 | 353 | 370 | 32 | 1436 |
| СЛОВО | 445 | 337 | 415 | 208 | 23 | 1428 |
| ПРОЦЕС |  | 20 | 152 | 1111 | 136 | 1419 |
| ПИТАННЯ | 43 | 74 | 521 | 283 | 477 | 1398 |
| УВЕСЬ | 254 | 455 | 403 | 173 | 110 | 1395 |
| МІСЦЕ | 223 | 202 | 330 | 240 | 380 | 1375 |
| **УКРАЇНСЬКИЙ** | 56 | 14 | 703 | 314 | 262 | 1349 |

*1*— belles-lettres*, 2* — colloquial*, 3* — journalistic*, 4* — scientific*, 5* — official genres.

The common lexical base size is 1389 words.

**The methodics of revealing for the conceptual model of the world**

It can be very important result if the selected list of words covers all the spheres of human activity. To verify this statement it would be good to have the conceptual or language model of the world. The conceptual model of the world, in our opinion, can be brought to light by comparing the ideographic dictionaries in different languages. Our hypothesis is the following: there is invariant synoptical scheme irrespective of language in all ideographic dictionaries. It is caused by the fact that human knowledge has the systematic nature, and language (in particular, the lexical composition) is its main holder, so they should be the similar system.

On this purpose we tried to collate the ideographic dictionaries synoptical schemes of English [*Roget* 1977], German [*Hallig & Wartburg* 1963; *Dornsief* 1963; *Meier* 1964], Spanish [*Casares* 1959], Czech [*Haller* 1974], Russian [*Morkovkin* 1984] and Ukrainian [*Sokolovska* 2002]. At first, we review very shortly those schemes without detailed description of their positive or negative sides aiming to show their general world-view differences. It is important to say that this is not linguistic but rather logical classification schemes of concepts.

**Roget's** International Thesaurus [*Roget* 1977] divided the English vocabulary on the first step into eight equipollent, as for him, groups with the next smaller subdivision: I "Abstract relations" (existence, relation, quantity, order, number, time, etc.), II "Space" (dimensions, structure, form, motion), III "Physics" (heat, light, electricity and electronics, mechanics, etc.), IV "Matter" (inorganic matter, organic matter), V "Sensation" (touch, taste, smell, sight, hearing, sound), VI "Intellect" (intellectual faculties and processes, state of mind, communication of ideas), VII "Volition" (condition, voluntary action, authority and control, support and opposition, possessive relations), VIII "Affection" (personal affections, sympathetic affections, morality, religion).

**F. Dornsief** [*Dornsief* 1963] divided the German vocabulary into 20 groups with the next smaller subdivision. The first and the second one cover the nature, which is understood here very widely: from cosmos, meteors, inorganic world through plants and animals to human body. The next six groups include the abstract and a priori concepts ("space", "size", "existence", "time", etc). The next four groups consist of the human psychological characteristics words: "wishes and actions", "sensation", "feeling, affects, feature of character", "though"). The words of four last groups describe the social relations and cultural phenomenon.

Another German language division proposed **R. Hallig and W. von Wartburg** [*Hallig & Wartburg* 1963]. They divided the universe into three main spheres: "universe", "human being", "human being and universe". Every of this spheres covers several conceptual field, and in the sum there are ten big complex fields ("heaven and heavenly body", "earth", "plant world", "animals world", "man as an alive being", "soul and mind", "man as a social being", "social organization and social institutions", "a priori", "science and technique"). Those fields have the next division.

The similar scheme lies in the basis of **Česky slovník věcný a synonymický** [*Haller* 1974]. The authors write in the preface that they depart from the R. Hallig and W. von Wartburg dictionary only in the case where Czech material needed another classification [*Haller* 1974, p. V]. In practice, the difference between the schemes of both dictionaries is minimum.

**H. Meier** [*Meier* 1964] has done the statistically based synopsis. He divided all the German vocabulary (11 million word occurrences) into 12 frequency zones: the first includes the most frequently words, the last includes the least frequently words. N. Karaulov said about an interesting fact of close result of two vocabulary classification (H. Meier and R. Hallig and W. von Wartburg) received by different methods [*Karaulov* 1967, p. 254].

**J. Casares** [*Casares* 1959] built his Spanish language dictionary scheme with God in the center. After God "universe" follows divided into inorganic ("matter and energy", "physics and chemistry", "geography, astronomy, meteorology", "geology, mineralogy") and organic matter (plant and animals). The animal world includes both "animal" and "man", the last group consist of "individual" and "society" with the following subdivision of individual into the groups "human as a living being", "human as an intellectual being" and "human as an agent of action", and the "society" divided into "communication, senses, thoughts", "social institutions", "work, service".

**A. Markowski** [*Markowski* 1990] created the scheme of Polish language with the word "I" on the top and three main fields: "I in the relations with myself" and "I in the relations with others" (with the relations with other people and other things). In the first field are: "I as a physical being" ("my body" and "something serving to my body") and "I as a psychic being" ("my thought" and "something serving to my thought"); in the second are: "I in the relations with God" ("my belief" i "something serving to my belief"), "I in the relations with people" ("my attitude to others" i "something serving to me and to others").

**V. Morkovkin** [*Morkovkin* 1984] proposed the hierarchic conceptual worldview of Russian language with regard to teaching methodics. In the "universe" on the base of dichotomous division he has divided conceptual spheres as follows: "abstract relation"–"material matter", "inorganic world"–"organic world", "plants"–"alive being", "unwise alive being"–"human being"; in "abstract relation" is separated general groups "existence", "space", "time", "changing", "quantity", "quality" and "relations".

Ukrainian scholar **Zh. Sokolowska** [*Sokolowska* 2002] has built the universal frame for any language (including Ukrainian) due to gnoseological and ontological parameters. The gnoseological concepts (cognition categories) such as existence, space, time, movement, something separate, quality, quantity, relation, are in the vertical column of table and ontological concepts (existence spheres), such as nature, man, society, are in the horizontal line. There are the words in the square where the lines cross.

After collating the ideographic dictionaries synoptical schemes of six different languages (see above) we can see in their center the common invariant part as follows: nature, including the spheres from heaven to animals, human being with the body and mental features, the relations between people in the society, and independent categories like existence, space, time, movement, etc.

**Specifics of the semantic structure of the Ukrainian language lexical base**

With the aim to find out what spheres of logically classified concept are covered by our existent words list, we should classify this list itself. In spite of different nationalities use the same scientific conceptual instrument, some concepts can have no separate lexemes for its notation in some languages, for instance, English *blue* – Ukrainian *synij, holubyj*; English *love,* Russian *ljubit'* – Ukrainian *ljubyty, koxaty*, etc. So, it can not have the equal classification and we do not agree with R. Tokarski equates the lexical and conceptual fields [*Tokarski* 1984, p. 11]. That is why we consider the semasiologic approaches to the vocabulary classification to be more natural for exact language classification because it is not fastened to the logical scheme for words but it goes from word to concept.

Our techniques of the language base classification was the following: on the first stage, parts of speech (as the most general linguo-philosophic categories) were selected. There were nine of them: noun, verb, adjective, pronoun, proverb, numeral, preposition, conjunction and particle. No interjection was found. There is also no article in Ukrainian.

On the second stage, basing on the common semantic features within parts of speech the words were joined into small groups (synonymic rows, antonymic pairs, hypero-hyponymical, partial-holonomy ("meronymy" in Lyon's term [*Malmkjær* 1991, p. 301]) and conversion-based groups). Different group types were found in different part of speech: synonymic and antonymic rows were found in all of them: synonymic (*šljax* 'way', *doroha* 'road'; *zaxodyty* 'to enter', *vxodyty* 'come into'; *tjažkyj* 'hard', *važkyj* 'difficult'; *viljnyj* 'free', *nezaležnyj* 'independent'; *zvyčajno* 'obviously', *očevydno* 'evidently'; *bilja* 'nearly', *poruč* 'close (to)'; *osj, ot* 'amplifier particle', etc.) and antonymic (*nadija* 'a hope', *strax* 'a fear'; *zaxodyty* 'to enter', *vyxodyty* 'to leave'; *xolodnyj* 'cold' – *harjačyj* 'hot'; *švydko* 'fast', *dovho* 'long'; *do* 'to', *vid* 'from'; *tak* 'yes', *ni* 'no'; *šče* 'yet', *vže* 'already', etc.).

Hypero-hyponymical groups were found in the nouns, verbs and adjectives (*kimnata* 'room' – *kabinet* 'cabinet', *klas* 'class', *zal* 'hall'; *počuvaty* 'to feel' – *ljubyty* 'to love'; *ljudsjkyj* 'human' – *žinočyj* 'feminine', etc.).

Conversion-based group are found in the nouns, verbs, prepositions and interjections (*čolovik* 'husband' – *družyna* 'wife', some nouns pairs of the model "*pryčyna* 'a cause; a reason' – *naslidok* 'effect'", *daty* 'to give' – *vzjaty* 'to take', *sered* 'in the middle' – *navkolo* 'round'; *jakščo* 'if' – *to* 'then' etc.).

And the partial-holonomy groups were found only in nouns (*tilo* 'body' – *holova* 'head', *ruka* 'hand', *noha* 'leg'; *ruka* 'hand' – *palecj* 'finger', etc.)

Then, on the third stage depending on the specifics of the semantic value of each word (denotative- or significative-based) these small groups were joined into lexical-semantic or thematic groups. The verbs create the lexical-semantic groups only, but the noun, pronoun and adverb have the lexical-semantic as well as thematic groups. For example, the nouns with denotative-based lexical meaning of natural formation create the thematic group corresponding to it: *hora* 'mounting', *pole* 'field', *lis* 'forest', *step* 'steppe', *more* 'see', *rička* 'river'. The nouns with significative-based lexical meaning of time create the lexical-semantic group: *čas* 'time', *rik* 'year', *misjacj* 'month', *tyždenj* 'week', *denj* 'day', *hodyna* 'hour', *xvylyna* 'minute', etc. The pronoun can be combined into lexical-semantic (e. g., "group of space": *korotkyj* 'short', *vysokyj* 'high', *nyzjkyj* 'low', *hlybokyj* 'deep') and thematic groups (e. g., "group of production": *vyrobnyčyj* 'production', *trudovyj* 'working', *robočyj* 'trade', *profesijnyj* 'professional', *texnologičnyj* 'technological') and so on. The lexical-semantic groups of time, movement, relation, space, etc. were distinguished in all the parts of speech.

Basing on these lexical-semantic groups in the case of verbs the lexical-semantic fields of movement, state, relation and others were distinguished. For nouns groups cannot be so strictly organized in such a strong fields. The most relevant differential features for noun meaning are: concrete / abstract. Within the concrete nouns the words were joined into animate / inanimate nature, human being and social relations. Within the abstract nouns the relevant feature was what kind of concept the word is connected: with a man, his work, mental or body characteristic, with nature or with abstract categories. We discovered the close situation in adjectives and in adverbs.

The last stage of lexical base classification the crystallization of general lexical fields covering all the parts of speech. There are fields of man, his body, mental features and mind, his work, individual relationships and attitude, social institutions and bureaucracy, animate and inanimate nature, general categories like time, space, existence, quality, quantity and some others. As we can observe, the word fields are quite correlative with conceptual groups from the invariant base of all the ideographic dictionaries.

But there are some distinctive features. For example, we can see the general tendency of lexical base abstractness. It became apparent not only in big number of abstract nouns, but in verbs general meaning as well. In many cases in lexical base is only the verb (the most neutral) naming the whole field or group in the ideographic dictionary (*hovoryty* 'to say' but not *šepeljavyty* 'burr', *kryčaty* 'cry', *šepotity* 'whisper', etc). There are big groups of words connecting with the norm (*typovyj* 'typical', *normaljnyj* 'normal', *normatyvnyj* 'normative', *vidpovidnyj* 'corresponding', *zvyčajnyj* 'usual', *pryrodnyj* 'natural', *osoblyvyj* 'special', etc.), working process (*stadija* 'stage', *etap* 'phase', *metod* 'method', *sposib* 'manner', *texnologija* 'technology', *pryjom* 'technique', *režym* 'procedure', etc.), leading profession (*kerivnyctvo* 'leadership', *prezydent* 'president', *dyrektor* 'director', *kerivnyk* 'chief', *zastupnyk* 'deputy director', etc.). We should take note of absence such groups as taste, sides of the world, seasons, days of the week. It is striking that there are *sjohodni* 'today', *zavtra* 'tomorrow' but no *včora* 'yesterday'; there is *dorohyj* 'expensive', but there is no *deševyj* 'cheep'; there is *žinočyj* 'feminine' but no *čolovičyj* 'masculine', there are *harjačyj* 'hot' and *xolodnyj* 'cold' but no *teplyj* 'warm'.

At this stage we can only establish the existence or absence of some of the words with some meanings, but the explanation of this phenomenon can be done only after future research. An accessary result of our analysis is the partial answer to the question "how language could be related to the world", considered by D. Geeraerts [*Aszer* 1994, p. 3804].

In spite of some indicated discrepancy, the list of words selected via formal techniques using the criteria of frequency and functional unrestrictedness covers practically all the conceptual filed. From this point of view, this list, being the lexical base of the Ukrainian language, might be called the compressed model of the world.